\documentclass[10pt,twocolumn,letterpaper]{article}

\usepackage[pagenumbers]{wacv}      

\usepackage{float}
\usepackage{multirow}
\usepackage{graphicx}
\usepackage{booktabs}

\definecolor{wacvblue}{rgb}{0.21,0.49,0.74}
\usepackage[pagebackref,breaklinks,colorlinks,allcolors=wacvblue]{hyperref}

\def\wacvPaperID{277} 
\def\confName{WACV}
\def\confYear{2027}

\title{ Semantic-Skeleton: Canonizing Quadruped Skeletons for Cross-Skeletal Pose Forecasting}

\author{
Aneeq Chowdhury$^{1}$ \quad
Ian Noronha$^{2}$\thanks{Jointly supervised this work.}  \thanks{Corresponding author. Email: \texttt{inoronha@purdue.edu}} \quad
Upinder Kaur$^{2}$\footnotemark[1] \\[3pt]
$^{1}$Department of Computer Science, Purdue University \\
$^{2}$Department of Agricultural and Biological Engineering, Purdue University \\
{\tt\small \{chowdh69, inoronha, kauru\}@purdue.edu}
}
\begin{document}
\maketitle
\begin{abstract}
Pose forecasting predicts future poses from observed motion, but existing methods typically assume fixed skeletal structures, limiting cross-species generalization. We propose Semantic-Skeleton, a graph-based framework for cross-skeletal 3D quadruped pose forecasting. Our method uses graph topology to map diverse quadruped skeletons into a shared canonical skeleton, allowing a single forecasting model to operate across inconsistent skeletal layouts. We pair this representation with a diffusion-based forecasting model to generate long-horizon motion predictions. We evaluate it on AnimalSyn3D across multiple forecasting horizons, motion subsets, and unseen-animal experiments. Across horizons, Semantic-Skeleton achieves an MPJPE of 0.05-0.08 meters and a PCP of 80\%-87\%. Compared with pose forecasting baselines evaluated under per-animal skeleton permutations, our method substantially improves both positional and structural accuracy, reducing MPJPE by 49.0\% on average and improving PCP by 20.8\%. In the unseen-animal setting, our method reduces MPJPE by 51.0\% on average, demonstrating stronger cross-skeleton generalization. We further validate Semantic-Skeleton on the real-world PFERD and AcinoSet datasets, where a shared cross-species model achieves performance comparable to species-specific models. These results show that structural canonization improves cross-skeletal forecasting, marking a step toward subject-agnostic animal pose forecasting across different skeletal structures. Code is available at \url{https://github.com/AneeqCML/Semantic-Skeleton.git}.
\end{abstract}

\section{Introduction}
\label{sec:intro}

Forecasting future poses from a sequence of past poses allows a system to anticipate how a subject will move. For animals, predicting motion is valuable for tracking, wildlife monitoring, and conservation, where changes in gait, posture, and behavior can reveal movement patterns and help anticipate future actions~\cite{thakkar2026forecasting,noronha2025mbeari,noronha2025quadforecaster}.

Pose forecasting builds directly on advances in pose estimation. Pose estimation has become an important tool for understanding subject motion by representing movement as a sequence of anatomical keypoints~\cite{zheng2020deep}. Early animal pose estimation focused on 2D keypoint tracking, with DeepLabCut~\cite{Mathis18DeepLabCut} and LEAP~\cite{Pereira19LEAP} bringing markerless 2D tracking to video at scale. SLEAP~\cite{Pereira22SLEAP} extended this to multi animal settings, and DANNCE~\cite{Karashchuk21DANNCE} and OpenMonkeyStudio~\cite{Bala20OpenMonkeyStudio} recovered 3D pose from multi camera capture. Large benchmarks such as AP-10K~\cite{Yu21AP10K} and APT-36K~\cite{yang2022apt} broadened 2D evaluation across species. 3D datasets including Animal3D~\cite{xu2023animal3d}, PFERD~\cite{li2024poses}, and AcinoSet~\cite{joska2021acinoset} added spatial structure, and synthetic resources such as DeformingThings4D~\cite{li20214dcomplete} and AnimalSyn3D~\cite{Fusco24ObjectAgnostic3D} supplied dense or temporally consistent motion. This progress has made reliable pose estimates available across a wide range of species, but turning those estimates into a forecast of future motion remains a challenge with sparse solutions.

While pose estimation is useful, pose forecasting further extends the study and behavioral modeling capabilities. However, pose forecasting has largely focused on humans. Early deep learning approaches used recurrent architectures such as ERD~\cite{Fragkiadaki15ERD}; later recurrent work~\cite{Martinez17HumanMotionPrediction} introduced residual RNN formulations that predict motion changes rather than absolute poses. Subsequent methods explored graph- and attention-based representations, including LearnTrajDep~\cite{wei2019motion}, HisRepItself~\cite{mao2020history}, and PGBIG~\cite{ma2022progressively}. Stochastic generative models have also been used to capture the multi-modal nature of future motion~\cite{aliakbarian2021contextually}, while diffusion-based methods such as DePOSit~\cite{saadatnejad2023diffusion} formulate forecasting as iterative denoising. However, these models generally assume a fixed skeleton, making them ill-suited for cross-species transfer where joint count, ordering, and anatomy vary.

Existing animal pose forecasting approaches are limited. QuadForecaster~\cite{noronha2025quadforecaster} trains species-specific diffusion models for 3D quadruped pose forecasting, while Thakkar et al.’s Forecasting Motion in the Wild~\cite{thakkar2026forecasting} explores dense point trajectories as a general motion representation across diverse non-rigid agents. These approaches either require separate species-specific models or avoid explicit 3D skeletal structure entirely, leaving open the problem of training a single skeleton-based model that generalizes across varying skeletons.

We study cross-skeleton quadruped pose forecasting: predicting future 3D poses for a quadruped, including animals unseen during training, using a model trained across quadrupeds with different joint counts, skeletal layouts, and keypoint definitions. This requires modeling both temporal motion and structural correspondence across skeletons.

We address this at the representation level. We introduce a canonization algorithm that aligns the skeletons of quadrupeds into a single shared joint space using graph topology. The algorithm selects a structural root for each skeleton, orders joints by graph level, and ranks joints within each level by structural importance, producing a fixed-size canonical pose that one forecasting model can consume regardless of an animal's native joint count or ordering. 
Our topology-based canonization relies on graph structure rather than consistent joint indices or species-specific anatomical labels (the joint index is used only as a final tie-breaker),
allowing the same procedure to be applied to quadrupeds with varying skeletons, and unseen quadrupeds. Using the resulting shared representation, we train a diffusion-based model for long-horizon 3D pose forecasting and compare it against learned models operating on native skeletons. We evaluate prediction accuracy across five forecasting horizons and multiple quadrupeds, including animals held out during training. We further test cross-skeletal robustness by deliberately permuting joint orderings and comparing permute-only and canonized preprocessing across five forecasting backbones, and validate the framework against native-skeleton baselines and on real-world AcinoSet and PFERD data. Overall, our contributions are:
\begin{itemize}
\item We formulate cross-skeleton quadruped pose forecasting, where a single model predicts future 3D poses across quadrupeds with different joint counts, skeletal layouts, and keypoint definitions, including quadrupeds unseen during training.

\item We introduce a graph-based canonization algorithm that maps different quadruped skeletons into a shared, structurally ordered joint space using graph topology rather than manual anatomical correspondence.

\item We demonstrate end-to-end cross-skeletal 3D pose forecasting by combining the canonical representation with a diffusion-based forecaster, evaluating across five forecasting horizons, high-motion settings, and quadrupeds held out during training.

\item We extensively validate the value of canonization through native-skeleton learned baselines and joint-permutation experiments across five forecasting architectures, showing that structural alignment improves cross-skeletal forecasting across different backbone designs.

\item We extend the framework to real-world PFERD and AcinoSet data, where a single shared model remains competitive with separately trained species-specific models despite operating across different skeletal structures.
\end{itemize}

\section{Related Work}
\label{sec:formatting}

\subsection{Animal Pose Estimation}
Animal pose estimation began with 2D markerless methods. DeepLabCut~\cite{Mathis18DeepLabCut} adapted pretrained human pose-estimation networks to track user-defined animal keypoints from limited labeled data. LEAP~\cite{Pereira19LEAP} similarly used supervised keypoint detection with iterative annotation, while SLEAP~\cite{Pereira22SLEAP} extended markerless pose estimation to multi-animal tracking using bottom-up and top-down approaches. These methods enabled practical behavioral analysis but operate primarily on frame-level 2D poses rather than forecasting future 3D motion across different skeletons.

Several datasets have driven progress in animal pose estimation. AP-10K~\cite{Yu21AP10K} provides over 10,000 annotated images across 54 mammal species, while APT-36K~\cite{yang2022apt} provides 36,000 annotated video frames across 30 species for pose estimation and tracking. Animal3D~\cite{xu2023animal3d} provides 3D pose and SMAL shape annotations for 40 mammal species, and AcinoSet~\cite{joska2021acinoset} contains synchronized multi-view recordings of free-running cheetahs with annotated dynamic locomotion. PFERD~\cite{li2024poses} contains synchronized multi-view recordings of diverse horse motions with dense full-body 3D marker annotations, along with fitted hSMAL pose representations. 
Synthetic datasets further expand supervision: DeformingThings4D~\cite{li20214dcomplete} provides large-scale 4D motion data, while AnimalSyn3D~\cite{Fusco24ObjectAgnostic3D} provides temporally consistent 3D skeleton labels across 13 animal categories for class-agnostic pose lifting and cross-rig generalization.
\subsection{Pose Forecasting}
Early pose forecasting relied on recurrent architectures. ERD\cite{Fragkiadaki15ERD} jointly learned pose representations and temporal dynamics using encoder-decoder recurrent networks, but later work showed that such models could suffer from short-term discontinuities and even underperform zero-velocity baselines. Martinez et al.~\cite{Martinez17HumanMotionPrediction} showed that a simpler residual sequence-to-sequence GRU with improved training could outperform more complex recurrent approaches.

Later methods increasingly exploited explicit spatial and temporal representations. LearnTrajDep~\cite{wei2019motion} represented joint trajectories with DCT coefficients and used a learned GCN to capture spatial dependencies. HisRepItself~\cite{mao2020history} combined historical motion attention with DCT-based GCN prediction, while PGBIG~\cite{ma2022progressively} progressively refined future poses using spatial and temporal GCNs. siMLPe~\cite{guo2022back} showed that lightweight MLPs with DCT encoding and residual displacement prediction could match or exceed more complex architectures. More recently, DePOSit~\cite{saadatnejad2023diffusion} formulated 3D pose forecasting as conditional denoising using cascaded short- and long-term diffusion models, supporting both clean and noisy observations.

Current animal pose forecasting remains limited. QuadForecaster~\cite{noronha2025quadforecaster} trained separate diffusion-based forecasting models on cheetah and cow pose sequences from the AcinoSet and MBE-ARI datasets, reporting results independently per species. Forecasting Motion in the Wild~\cite{thakkar2026forecasting} moves away from skeletal pose entirely, representing animal behavior as dense 2D point trajectories and training a diffusion transformer to forecast unordered sets of future point tracks while handling occlusion. However, these methods either rely on species-specific skeletal models or avoid 3D skeletal pose altogether, leaving unified 3D pose forecasting across different animal skeletons unaddressed.

\section{Methods}

\subsection{Defining a Semantic Joint Ordering Across Different Quadrupeds}
The quadrupeds in our dataset have varying skeletal structures, with different numbers of joints and inconsistent joint indices across species. As a result, the same joint index may correspond to different anatomical locations across animals, making direct cross-species forecasting difficult.
\subsubsection{Topology-Based Root Selection}
We represent each animal skeleton as an undirected graph, where vertices correspond to joints and edges correspond to skeletal connections. For each quadruped, we select the highest-degree vertex as the root joint, defined as the joint with the largest number of direct skeletal connections. This root serves as a shared structural anchor from which the remaining joints are organized during canonization.
If several joints share the highest degree, the one with the lowest joint index is chosen.
\subsubsection{Levels Traversal}
We perform a standard levels traversal of the skeleton, starting from the root, and label each joint by its level, with the root being level 0, and the furthest joint, or leaf vertices being level 6. 
\subsubsection{Level-Based Joint Quotas and Tie-Breaking}
Within each BFS level, joints are ranked by structural importance using subtree size followed by node degree. Remaining ties are resolved using mean distance to leaf joints, followed by geometry and motion-based cues. A body-centric frame is estimated for each sequence using velocity for the forward/backward axis and pose PCA for lateral and vertical directions. Joints are projected onto this frame to obtain front/back, left/right, and up/down scores, with the original joint index as the final deterministic tie-breaker.

We retain a fixed number of joints per BFS level, based on the maximum shared across subject quadrupeds and adjusted for other datasets. The seven-level quotas are:

\[
\{0:1,\; 1:6,\; 2:3,\; 3:5,\; 4:4,\; 5:4,\; 6:2\}.
\]

This converts skeletons with different joint counts into a consistent canonical representation while preserving structurally important joints.
\subsubsection{Canonical Pose Representation and Root Normalization} After canonization, each quadruped pose is represented using a fixed set of $J=25$ joints. Each joint contains its 3D spatial coordinates, resulting in a pose representation $X_t \in \mathbb{R}^{J \times 3}$ at timestep $t$. We apply root normalization by subtracting the root joint position from every selected joint at each timestep, producing a root-relative pose representation. This removes global translation and allows the model to focus on relative skeletal configuration and motion.
The resulting representation is not the original dataset joint indices. Instead, 
each canonical index corresponds to a slot defined by graph level, structural priority, and the per-level quota. Slots are thus consistent in structural rank across skeletons, though they are not guaranteed to correspond to the same anatomical limb in every animal. Thus, the first joint corresponds to the selected root, while subsequent joints correspond to the highest-ranked joints within each BFS level according to the canonical ordering procedure.

\subsection{Diffusion Model}

After constructing the canonical pose representation, we forecast future motion using a conditional diffusion model. Given an input motion history of $N$ frames and a prediction horizon of $M$ frames, each pose is represented as $X_t \in \mathbb{R}^{J \times 3}$, where $J=25$. We flatten each pose into a vector in $\mathbb{R}^{3J}$ and represent the full motion window as
$X_0 \in \mathbb{R}^{(N+M)\times 3J}$.
A binary conditioning mask $m$ is used to distinguish observed and target frames, where $m=1$ for the observed input history and $m=0$ for the future frames to be predicted.

We formulate pose forecasting as CSDI-style conditional diffusion-based imputation~\cite{tashiro2021csdi}. During training, Gaussian noise is added to the motion window according to a diffusion process:
\[
X^{(\tau)} = \sqrt{\bar{\alpha}_\tau}X_0 + \sqrt{1-\bar{\alpha}_\tau}\epsilon,
\]
where $\epsilon \sim \mathcal{N}(0,I)$, $\alpha_\tau = 1-\beta_\tau$, and $\bar{\alpha}_\tau = \prod_{s=1}^{\tau}\alpha_s$. We use a cosine noise schedule with $50$ diffusion steps.

The denoising network $\epsilon_\theta$ is trained to predict the added noise from the noisy motion, the diffusion timestep, and the conditioning information:
$\hat{\epsilon} = \epsilon_\theta(X^{(\tau)}, \tau, m, X_{\text{obs}})$.
The conditioning information includes the observed pose history, a temporal embedding for each frame, a feature embedding for each pose coordinate, and the binary observation mask. This allows the model to preserve the known input frames while generating the masked future sequence.

The training objective is the mean squared error between the true noise and the predicted noise, evaluated only over the masked future region:
\[
\mathcal{L}_{\text{diff}} =
\frac{
\left\|
(\epsilon - \epsilon_\theta(X^{(\tau)},\tau,m,X_{\text{obs}})) \odot (1-m)
\right\|_2^2
}{
\sum (1-m)
}.
\]
This encourages the model to learn the distribution of plausible future poses conditioned on the observed motion history, rather than reconstructing the already observed frames.

At inference time, the future region is initialized with Gaussian noise and iteratively denoised through the learned reverse diffusion process. The observed input frames remain fixed throughout sampling, while only the masked future frames are generated. The final output is a forecasted sequence of canonical, root-normalized 3D poses in the shared quadruped pose space.

During training, the model is optimized on canonized pose windows sampled from the training set. We use Adam with a learning rate of $10^{-3}$ and weight decay of $10^{-6}$. The learning rate is decayed using a multi-step schedule at $75\%$ and $90\%$ of training. 

\subsubsection{Short- and Long-Horizon Forecasting} We train diffusion models for both short- and long-horizon forecasting. For short-horizon prediction, a single model predicts the next few future frames directly from the observed sequence. For longer horizons, we use a two-stage strategy: a short-horizon model first predicts an intermediate future segment, and the mean prediction across $N$ diffusion samples is inserted back into the sequence as additional observed context. A second long-horizon model then forecasts the remaining future motion, again using the mean across $N$ generated samples as the final prediction. This allows long-horizon forecasting to condition on both the original observed history and an estimated near-future trajectory.

\begin{figure*}[t]
    \centering
    \includegraphics[page=1,width=0.98\textwidth]{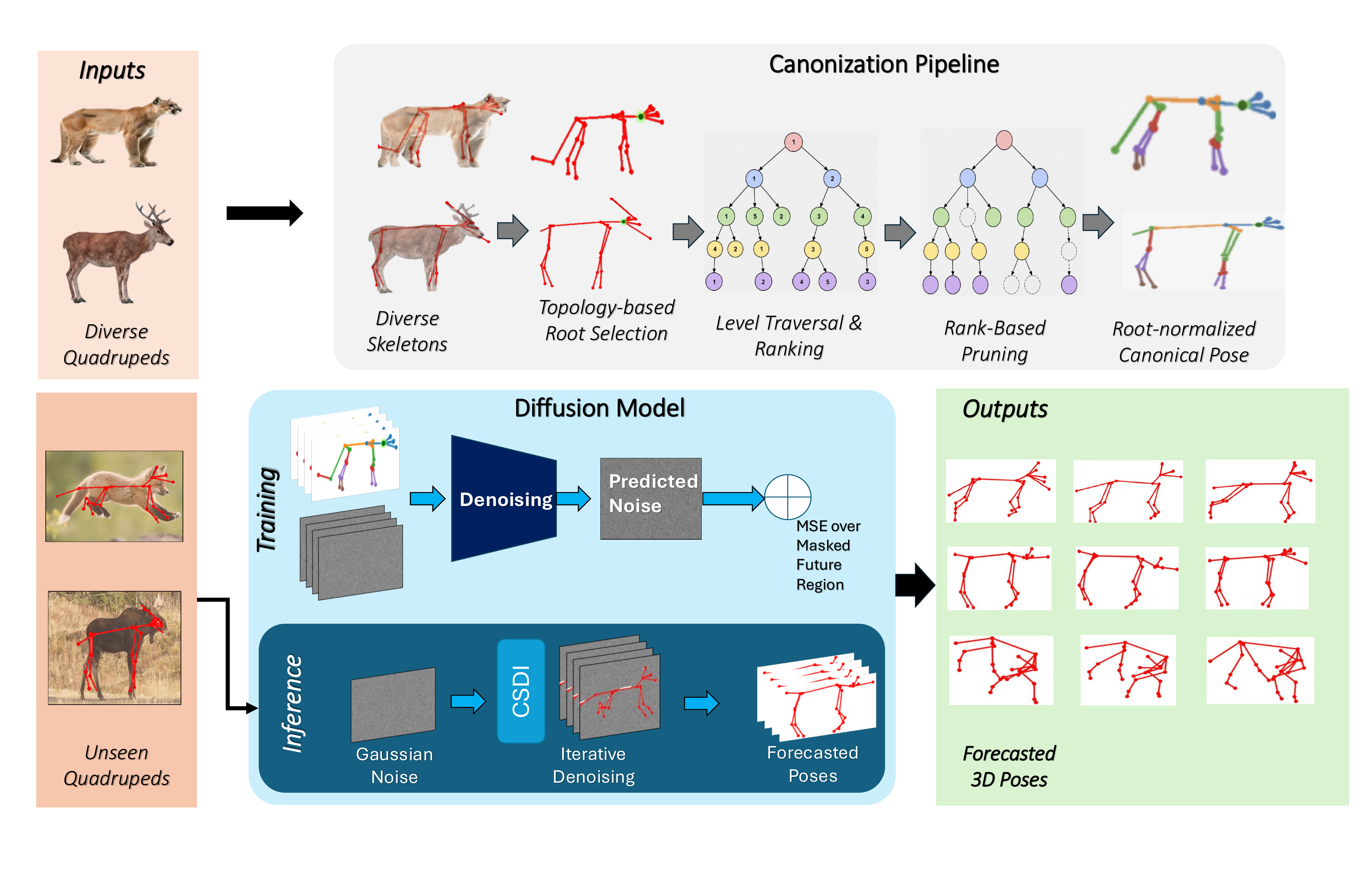}\vspace{-1.0em}
    \caption{Overview of the Semantic-Skeleton framework. Diverse quadruped skeletons are first canonized through topology-based root selection, breadth-first level traversal, and structural ranking. This produces a root-normalized canonical pose shared across species. A diffusion model is then trained to denoise masked future frames within this canonical space, supervised by an MSE loss computed only over the masked region. At inference, the same canonical representation and iterative denoising procedure forecast future poses for quadrupeds unseen during training.}\vspace{-1.0em}
    \label{fig:slide1}
\end{figure*}

\section{Experiments}
We evaluate cross-skeleton quadruped pose forecasting across animals with different skeletal structures, testing whether canonization improves accuracy, structural consistency, and unseen-animal generalization across backbones, high-motion settings, and ablations. For consistency within AnimalSyn3D, quadrupeds with fewer than 25 joints, unlike the majority, were excluded to improve generalization.
\subsection{Experimental Setup}
Additionally, we use the following metrics: MPJPE\cite{Calabrese_2019_CVPR_Workshops}, which measures the mean joint position error between the predicted and ground-truth poses; PCP\cite{SARAFIANOS20161}, which measures the percentage of correctly predicted skeletal parts; Bone Error\cite{8237546}, which measures edge-level positional accuracy by averaging the Euclidean errors of the two joints defining each skeletal bone; Bone Length Error\cite{tchenegnon2025mocosys} which measures the mean absolute difference between predicted and ground-truth bone lengths; and BoneStd\cite{8237546}, which measures the temporal variation of predicted bone lengths across the forecast horizon. For all PCP calculations, we used $\alpha = 0.5$, and all MPJPE results are reported in meters. 
Let $\mathcal{E}$ denote the set of skeleton edges, $T$ denote the number of forecasted time steps, and $\mathbf{x}_{t,j}$ and $\hat{\mathbf{x}}_{t,j}$ the ground-truth and predicted 3D positions of joint $j$ at time $t$. For each edge
$(s,e)\in\mathcal{E}$ and time step $t$, define
\[
\begin{aligned}
\epsilon^s_t &= \|\hat{\mathbf{x}}_{t,s}-\mathbf{x}_{t,s}\|_2, &
\epsilon^e_t &= \|\hat{\mathbf{x}}_{t,e}-\mathbf{x}_{t,e}\|_2,\\
\hat{d}_{t,s,e} &= \|\hat{\mathbf{x}}_{t,s}-\hat{\mathbf{x}}_{t,e}\|_2, &
d_{t,s,e} &= \|\mathbf{x}_{t,s}-\mathbf{x}_{t,e}\|_2,\\
\bar{d}_{s,e} &= \frac{1}{T}\sum_{t=1}^{T}\hat{d}_{t,s,e}.
\end{aligned}
\]
The bone-based metrics are:
\[
\begin{aligned}
\mathrm{BoneErr}
&=
\frac{1}{T|\mathcal{E}|}
\sum_{t=1}^{T}
\sum_{(s,e)\in\mathcal{E}}
\frac{\epsilon^s_t+\epsilon^e_t}{2},\\
\mathrm{PCP}
&=
\frac{1}{T|\mathcal{E}|}
\sum_{t=1}^{T}
\sum_{(s,e)\in\mathcal{E}}
\mathbf{1}\!\left[
\frac{\epsilon^s_t+\epsilon^e_t}{2}
\leq
\alpha d_{t,s,e}
\right],\\
\mathrm{BoneLenErr}
&=
\frac{1}{T|\mathcal{E}|}
\sum_{t=1}^{T}
\sum_{(s,e)\in\mathcal{E}}
\left|\hat{d}_{t,s,e}-d_{t,s,e}\right|,\\
\mathrm{BoneStd}
&=
\frac{1}{|\mathcal{E}|}
\sum_{(s,e)\in\mathcal{E}}
\sqrt{
\frac{1}{T}
\sum_{t=1}^{T}
\left(\hat{d}_{t,s,e}-\bar{d}_{s,e}\right)^2
}.
\end{aligned}
\]
\subsection{Canonical Pose Forecasting with Diffusion}
We evaluate the end-to-end framework by applying canonization followed by diffusion-based forecasting. The dataset is split approximately 79/21\% for training and evaluation, with one animal held out entirely for zero-shot evaluation. The data span diverse animal morphologies and actions, while a separate high-motion evaluation with large pose changes tests generalization beyond low-motion sequences.

All sequences are canonized for training, and predictions are evaluated in the canonical, root-relative joint space. Bone-based metrics use the skeleton connections between the joints retained for each animal after canonization.
We consider five forecasting horizons: $20\rightarrow15$, $20\rightarrow20$, $20\rightarrow25$, $25\rightarrow15$, and $25\rightarrow20$. Short-horizon $20\rightarrow5$ and $25\rightarrow5$ models provide intermediate context to the corresponding long-horizon models. At inference, the short prediction is appended as context, and the mean of five diffusion samples is used as the final prediction.

We report MPJPE, First-5 MPJPE, PCP, Bone Error, Bone Length Error, and BoneStd. First-5 MPJPE measures the short-horizon prediction, while MPJPE measures the remaining long-horizon forecast. We additionally compare against zero-velocity and constant-velocity baselines. Zero velocity repeats the last observed pose across the forecast horizon, while constant velocity extrapolates the displacement between the final two observed frames, testing whether performance can be explained by static or simple linear motion. We also train and compare to learned baselines on animals with identical joint counts, since current architectures  require fixed input/output dimensions. Semantic-Skeleton was trained across animals while the learned baselines were species specific.

\begin{table}[t]
\centering
\caption{
Forecasting performance across input-output horizons. Lower is better except for PCP.
}\vspace{-1.0em}
\label{tab:horizon_forecasting_results}
\renewcommand{\arraystretch}{1.15}
\setlength{\tabcolsep}{1.3pt}
\scriptsize
\begin{tabular}{l c c c c c}
\toprule
\textbf{Horizon}
& \textbf{MPJPE $\downarrow$}
& \textbf{First-5 MPJPE $\downarrow$}
& \textbf{Zero MPJPE $\downarrow$}
& \textbf{Const. Vel. $\downarrow$}
& \textbf{PCP $\uparrow$} \\
\midrule
$20 \rightarrow 15$ & 0.0527 & 0.0121 & 0.0894 & 0.0810 & 0.8552 \\
$20 \rightarrow 20$ & 0.0695 & 0.0119 & 0.1029 & 0.1080 & 0.8108 \\
$20 \rightarrow 25$ & 0.0802 & 0.0114 & 0.1111 & 0.1327 & 0.7971 \\
$25 \rightarrow 15$ & 0.0535 & 0.0121 & 0.0834 & 0.0765 & 0.8617 \\
$25 \rightarrow 20$ & 0.0634 & 0.0119 & 0.0951 & 0.1007 & 0.8386 \\
\bottomrule
\end{tabular}\vspace{-1.0em}
\end{table}

\begin{table}[t]
\centering
\caption{
Forecasting performance across learned backbones and native skeleton sizes. Lower MPJPE and higher PCP are better.
}
\vspace{-1.0em}
\label{tab:backbone_forecasting_results}
\renewcommand{\arraystretch}{1.02}
\setlength{\tabcolsep}{2.5pt}
\scriptsize

\begin{tabular}{c l c cc}
\toprule
\textbf{Joints}
& \textbf{Backbone}
& \textbf{Horizon}
& \textbf{MPJPE $\downarrow$}
& \textbf{PCP $\uparrow$} \\
\midrule

\multirow{5}{*}{26}
& PGBIG             & $10\!\to\!15$ & 0.051 & 0.790 \\
& HisRepItself      & $25\!\to\!15$ & 0.101 & 0.695 \\
& LearnTrajDep      & $20/25\!\to\!15$ & 0.054 / 0.050 & 0.831 / 0.830 \\
& siMLPe            & $20/25\!\to\!15$ & 0.054 / 0.052 & 0.800 / 0.795 \\
& Semantic-Skeleton & $20/25\!\to\!15$ & \textbf{0.043 / 0.038} & \textbf{0.886 / 0.905} \\

\midrule

\multirow{5}{*}{28}
& PGBIG             & $10\!\to\!15$ & 0.030 & 0.827 \\
& HisRepItself      & $25\!\to\!15$ & 0.049 & 0.717 \\
& LearnTrajDep      & $20/25\!\to\!15$ & 0.044 / 0.039 & 0.714 / 0.740 \\
& siMLPe            & $20/25\!\to\!15$ & 0.031 / 0.023 & 0.826 / 0.882 \\
& Semantic-Skeleton & $20/25\!\to\!15$ & \textbf{0.023 / 0.021} & \textbf{0.891 / 0.902} \\

\midrule

\multirow{5}{*}{29}
& PGBIG             & $10\!\to\!15$ & 0.102 & 0.834 \\
& HisRepItself      & $25\!\to\!15$ & 0.072 & 0.916 \\
& LearnTrajDep      & $20/25\!\to\!15$ & 0.114 / 0.107 & 0.815 / 0.830 \\
& siMLPe            & $20/25\!\to\!15$ & 0.083 / 0.088 & 0.870 / 0.867 \\
& Semantic-Skeleton & $20/25\!\to\!15$ & \textbf{0.041 / 0.032} & \textbf{0.951 / 0.966} \\

\bottomrule
\end{tabular}
\end{table}
\begin{table}[t]
\centering
\caption{
Forecasting performance across input-output horizons for the unseen animal Fox. Lower is better except for PCP.
}\vspace{-1.0em}
\label{tab:horizon_forecasting_results_unseen_fox}
\renewcommand{\arraystretch}{1.15}
\setlength{\tabcolsep}{1.3pt}
\footnotesize
\begin{tabular}{l c c c c c}
\toprule
\textbf{Horizon}
& \textbf{MPJPE $\downarrow$}
& \textbf{First-5 MPJPE $\downarrow$}
& \textbf{Zero MPJPE $\downarrow$}
& \textbf{Const. Vel. $\downarrow$}
& \textbf{PCP $\uparrow$} \\
\midrule
$20 \rightarrow 15$ & 0.0148 & 0.0041 & 0.0198 & 0.0190 & 0.8757 \\
$20 \rightarrow 20$ & 0.0161 & 0.0035 & 0.0191 & 0.0216 & 0.8688 \\
$20 \rightarrow 25$ & 0.0144 & 0.0029 & 0.0153 & 0.0194 & 0.8919 \\
$25 \rightarrow 15$ & 0.0133 & 0.0044 & 0.0173 & 0.0161 & 0.8988 \\
$25 \rightarrow 20$ & 0.0129 & 0.0036 & 0.0146 & 0.0155 & 0.9041 \\
\bottomrule
\end{tabular}
\end{table}
Across all horizons, Semantic-Skeleton achieves an MPJPE of 0.05--0.08 m and PCP of 0.80--0.87. Overall forecasting results are reported in Table~\ref{tab:horizon_forecasting_results}, with unseen-animal and high-motion results reported separately. For high-motion evaluation in Table ~\ref{tab:horizon_high_motion_85}, motion is measured as the average per-joint frame-to-frame displacement, and we report windows between the 85th and 95th percentiles. On a Tiger, Semantic-Skeleton achieves MPJPE and PCP of 0.121020/0.804882, and Raccoon 0.045980/0.761455.

Table~\ref{tab:backbone_forecasting_results} compares Semantic-Skeleton with learned species-specific models operating directly on native 26, 28, and 29-joint skeletons. Semantic-Skeleton remains competitive across all skeleton sizes and both horizons, and in all cases achieves the best MPJPE and PCP. These values differ from Table~\ref{tab:horizon_forecasting_results} because the backbone comparison evaluates the full 15-frame forecast horizon. Overall, the results show that semantic canonization supports heterogeneous skeletal structures while remaining competitive with models trained directly on fixed native skeletons.

\begin{table}[t]
\centering
\caption{
High-motion forecasting results for the 85th--95th percentile motion bucket. Lower is better except for PCP.
}\vspace{-1.0em}
\label{tab:horizon_high_motion_85}
\renewcommand{\arraystretch}{1.15}
\setlength{\tabcolsep}{13pt}
\footnotesize
\begin{tabular}{l c c c}
\toprule
\textbf{Horizon}
& \textbf{MPJPE $\downarrow$}
& \textbf{Bone Error $\downarrow$}
& \textbf{PCP $\uparrow$} \\
\midrule
$20 \rightarrow 15$ & 0.0962 & 0.0851 & 0.8215 \\
$20 \rightarrow 20$ & 0.1317 & 0.1162 & 0.7243 \\
$20 \rightarrow 25$ & 0.1555 & 0.1389 & 0.6594 \\
$25 \rightarrow 15$ & 0.0974 & 0.0864 & 0.8157 \\
$25 \rightarrow 20$ & 0.1278 & 0.1134 & 0.7329 \\
\bottomrule
\end{tabular}
\end{table}
\begin{table}[t]
\centering
\caption{
Structural forecasting metrics across input-output horizons. Lower is better.
}\vspace{-1.0em}
\label{tab:horizon_structural_results}
\renewcommand{\arraystretch}{1.15}
\setlength{\tabcolsep}{10pt}
\footnotesize
\begin{tabular}{l c c c}
\toprule
\textbf{Horizon}
& \textbf{Bone Error $\downarrow$}
& \textbf{Bone Length $\downarrow$}
& \textbf{Bone Std. $\downarrow$} \\
\midrule
$20 \rightarrow 15$ & 0.0467 & 0.0081 & 0.0031 \\
$20 \rightarrow 20$ & 0.0620 & 0.0106 & 0.0044 \\
$20 \rightarrow 25$ & 0.0719 & 0.0114 & 0.0055 \\
$25 \rightarrow 15$ & 0.0474 & 0.0079 & 0.0032 \\
$25 \rightarrow 20$ & 0.0563 & 0.0092 & 0.0044 \\
\bottomrule
\end{tabular}
\end{table}

\subsection{Forecasting on Real-World Data}
In this experiment, we evaluate whether our canonization procedure enables a single forecasting model to learn across quadrupeds from real-world 3D pose datasets. Because real-world 3D quadruped pose datasets are scarce, we use PFERD and AcinoSet; Animal3D is unsuitable due to limited data and non-temporal sequences.

Before training, we preprocess both datasets for consistency. We downsample PFERD and a subset of AcinoSet sequences recorded at higher than 90 FPS to 90 FPS and balance the training data by selecting approximately equal numbers of frames from each dataset.

PFERD contains 36 joints and a deeper kinematic graph, whereas AcinoSet contains 20. We therefore coarsen
PFERD level-by-level by removing intermediate joints along densely sampled kinematic chains while retaining major structural landmarks, producing a 21-joint skeleton with graph depth more comparable to AcinoSet. We then apply the same canonization procedure to both datasets. For PFERD, BFS levels are computed on the simplified graph, while degree and subtree size are measured on the original 36-joint topology; for AcinoSet, all features are computed on its native skeleton. Because these skeletons contain fewer BFS levels than our synthetic setting, we use the canonical level quota \{0:1,\; 1:4,\; 2:6,\; 3:7,\; 4:2\}, selected from the maximum number of joints at each BFS level shared across the two skeletons to yield a shared 20-joint representation. More generally, this quota can be adapted to other datasets based on the maximum number of joints shared across subjects at each BFS level, as discussed above. 
As dataset-specific baselines, we train separate diffusion models on each dataset's native skeleton using the same forecasting architecture. A native cross-skeleton baseline is not directly applicable because standard forecasting models require fixed input and output dimensions, while the datasets have different joint counts. All evaluation is done on a $20\rightarrow15$ horizon.

Semantic-Skeleton performs comparably to the dataset-specific models and in several cases matches or exceeds them despite using a single shared model. On AcinoSet, it improves MPJPE from 0.0784 to 0.0765 and PCP from 0.8235 to 0.8331; on PFERD, it substantially improves PCP from 0.6667 to 0.8327. These results suggest that the shared representation can generalize across different quadruped skeletons while remaining competitive with models trained separately on each dataset, as shown in Table \ref{tab:real_data}.
\begin{table}[t]
\centering
\small
\setlength{\tabcolsep}{5pt}

\begin{tabular}{lcccc}
\hline
& \multicolumn{2}{c}{\textbf{AcinoSet}} 
& \multicolumn{2}{c}{\textbf{PFERD}} \\
\hline
\textbf{Model} 
& \textbf{MPJPE} & \textbf{PCP}
& \textbf{MPJPE} & \textbf{PCP} \\
\hline
AcinoSet Model      & 0.0784 & 0.8235 & -- & -- \\
PFERD Model         & -- & -- & 0.1035 & 0.6667 \\
Semantic-Skeleton   & 0.0765 & 0.8331 & 0.1121 & 0.8327 \\
\hline
\end{tabular}

\caption{Comparison of species-specific AcinoSet and PFERD models with Semantic-Skeleton on real-world data.}
\label{tab:real_data}
\end{table}

\subsection{Canonization Across Forecasting Backbones}
To evaluate whether canonization improves performance across forecasting architectures, we compare DePOSit, LearnTrajDep, siMLPe, PGBIG, and HisRepItself under permute-only and canonized preprocessing. Because standard forecasting models require fixed input/output dimensions, they cannot be trained jointly across animals with different native joint counts. We therefore use animals with 26 joints and canonize them to a shared 25-joint representation. In the permute-only setting, joint order is fixed within each animal but differs across animals, allowing us to isolate the effect of semantic alignment. Skeleton edges are mapped to the corresponding permuted or canonized spaces before computing metrics.
Most models use a $20\rightarrow15$ horizon; HisRepItself uses $25\rightarrow15$ and PGBIG uses $10\rightarrow15$ for architectural compatibility. Results are reported on the full evaluation set and an unseen Elk split, which is separate from the Fox split used in Section~4.2.
As shown in Tables~\ref{tab:permute_vs_canonical_delta} and~\ref{tab:unseen_elk_permute_vs_canonical}, canonization consistently reduces MPJPE and Bone Error while increasing PCP across all five backbones, with similar gains on the unseen animal.

\begin{table}[t]
\centering
\caption{
Permute-only versus canonized preprocessing across forecasting models. Lower is better for MPJPE and Bone Error; higher is better for PCP.
}\vspace{-1.0em}
\label{tab:permute_vs_canonical_delta}
\resizebox{\columnwidth}{!}{
\begin{tabular}{l ccc ccc}
\toprule
& \multicolumn{3}{c}{\textbf{Permute-only}} & \multicolumn{3}{c}{\textbf{Canonized}} \\
\cmidrule(lr){2-4} \cmidrule(lr){5-7}
\textbf{Model} & MPJPE $\downarrow$ & Bone Err $\downarrow$ & PCP $\uparrow$ & MPJPE $\downarrow$ & Bone Err $\downarrow$ & PCP $\uparrow$ \\
\midrule
DePOSit      & 0.5475 & 0.5420 & 0.6544 & 0.1116 & 0.1037 & 0.8238 \\
LearnTrajDep & 0.1959 & 0.1951 & 0.6218 & 0.0942 & 0.0835 & 0.8515 \\
siMLPe       & 0.1613 & 0.1551 & 0.7514 & 0.0861 & 0.0764 & 0.8648 \\
PGBIG        & 0.1523 & 0.1490 & 0.7373 & 0.1220 & 0.1066 & 0.7842 \\
HisRepItself & 0.2342 & 0.2288 & 0.6582 & 0.1245 & 0.1116 & 0.7865 \\
\bottomrule
\end{tabular}
}\vspace{-1.0em}
\end{table}

\begin{table}[t]
\centering
\caption{
Permute-only vs. canonized preprocessing on the unseen animal subset (Elk). 
}\vspace{-1.0em}
\label{tab:unseen_elk_permute_vs_canonical}
\resizebox{\columnwidth}{!}{
\begin{tabular}{l ccc ccc}
\toprule
& \multicolumn{3}{c}{\textbf{Permute-only}} & \multicolumn{3}{c}{\textbf{Canonized}} \\
\cmidrule(lr){2-4} \cmidrule(lr){5-7}
\textbf{Model} & MPJPE $\downarrow$ & Bone Err $\downarrow$ & PCP $\uparrow$ & MPJPE $\downarrow$ & Bone Err $\downarrow$ & PCP $\uparrow$ \\
\midrule
\multicolumn{7}{c}{\textbf{Unseen Animal: Elk}} \\
\midrule
DePOSit      & 1.6589 & 1.6691 & 0.0867 & 0.1224 & 0.1091 & 0.8211 \\
LearnTrajDep & 0.2302 & 0.2309 & 0.5693 & 0.1014 & 0.0898 & 0.8521 \\
siMLPe       & 0.1814 & 0.1747 & 0.7371 & 0.0936 & 0.0829 & 0.8642 \\
PGBIG        & 0.1713 & 0.1680 & 0.7216 & 0.1302 & 0.1135 & 0.7840 \\
HisRepItself & 0.2034 & 0.1960 &  0.6898 & 0.1338 & 0.1197 & 0.7855 \\
\bottomrule
\end{tabular}
}\\
\footnotesize{Lower is better for MPJPE and Bone Error; higher is better for PCP.}\vspace{-2.0em}
\end{table}
The gains are especially large for DePOSit, improving from 0.5475 to 0.1116 MPJPE and 0.6544 to 0.8238 PCP on the full evaluation set, and from 1.6589 to 0.1224 MPJPE and 0.0867 to 0.8211 PCP on unseen Elk. These results show that semantic alignment substantially improves cross-skeletal generalization, providing a transferable representation across skeletons.
\begin{figure}[t]
    \centering
    \includegraphics[width=0.98\linewidth]{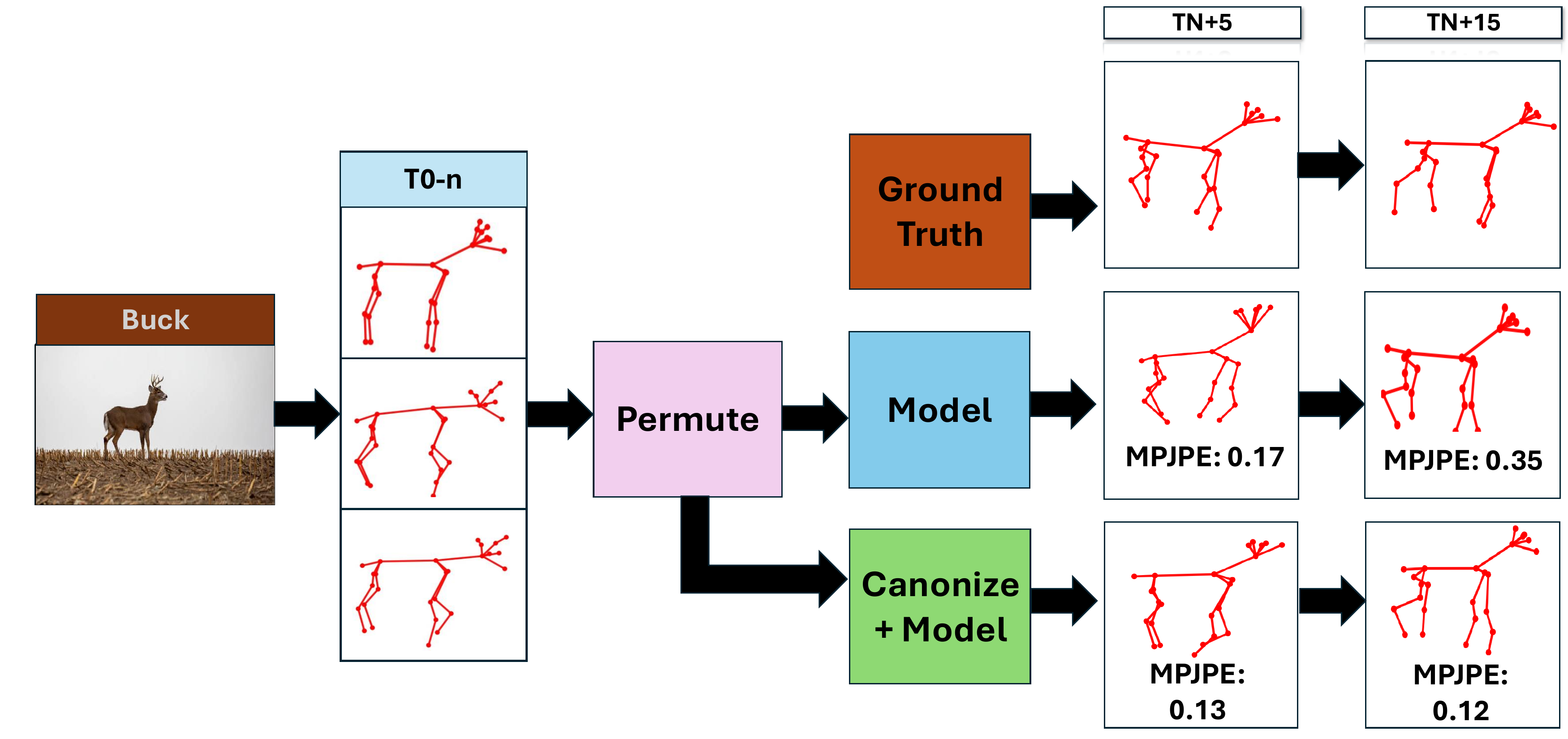}\vspace{-1.0em}
    \caption{Qualitative comparison of permute-only and canonized preprocessing for a buck sequence.}\vspace{-1.0em}
    \label{fig:buck_comparison}
\end{figure}

\subsection{Gait-Specific Forecasting Analysis}

To examine performance across motion patterns, we evaluate Semantic-Skeleton on representative actions spanning low-motion, locomotion, and dynamic behaviors. As shown in Table~\ref{tab:gait_action_results}, idle is easiest to predict (MPJPE = 0.0121) due to limited motion, while jumping is most challenging (0.0559), potentially due to abrupt transitions between takeoff, flight, and landing. Walking (0.0442) and trotting (0.0379) fall between these extremes, likely benefiting from more regular motion patterns. Overall, these results suggest that motion complexity and temporal regularity influence forecasting difficulty.
\begin{table}[H]
\centering
\caption{
Forecasting performance across representative gait and action categories for the $20\!\rightarrow\!15$ setting. Lower MPJPE and higher PCP are better.
}
\vspace{-1.0em}
\label{tab:gait_action_results}
\renewcommand{\arraystretch}{1.10}
\setlength{\tabcolsep}{5pt}
\footnotesize

\begin{tabular}{lcccccc}
\toprule
\textbf{Metric}
& \textbf{Idle}
& \textbf{Walk}
& \textbf{Trot}
& \textbf{Jump}
& \textbf{Swim}
& \textbf{Climb} \\
\midrule
MPJPE $\downarrow$ & 0.0121 & 0.0442 & 0.0379 & 0.0559 & 0.0396 & 0.0284 \\
PCP $\uparrow$     & 0.9830 & 0.8694 & 0.8518 & 0.8037 & 0.8934 & 0.8606 \\
\bottomrule
\end{tabular}

\end{table}
\subsection{Ablation Study}

In this experiment, we isolate the effects of graph-based joint ordering, deterministic within-level selection, and root normalization. \textbf{Canonical} applies our BFS-based topology canonization to a shared 25-joint representation.

\textbf{Canonical + no root normalization} removes the root-relative coordinate normalization, while \textbf{Canonical + random within levels} retains BFS-level grouping but randomizes joint selection within each level.

As non-canonical baselines, \textbf{Random permutation per animal} independently permutes each animal's joints and retains the first 25 without graph-based alignment, testing whether matching joint count alone is sufficient. Its no-root-normalization variant further removes the shared root-relative coordinate frame.

Finally, \textbf{Random permutation per animal $\rightarrow$ canonical} permutes each skeleton and remaps its edges before canonization, testing whether structure can be recovered independently of the original joint indices. The corresponding random-within-level variant tests whether this recovery depends on deterministic within-level ordering.

\begin{table}[t]
\centering
\caption{
  Ablation: Canonical Ordering and Root Normalization
}\vspace{-1.0em}
\label{tab:ablation_study}
\footnotesize
\setlength{\tabcolsep}{4pt}
\begin{tabular}{@{}l c c c c@{}}
  \toprule
  \textbf{Ordering} & \textbf{RN} & \textbf{MPJPE$\downarrow$} & \textbf{BE$\downarrow$} & \textbf{PCP$\uparrow$} \\
  \midrule
  Can.                         & \checkmark & 0.0940 & 0.0840 & 0.8413 \\
  Can.                         & $\times$   & 0.7790 & 0.7648 & 0.6470 \\
  Can. + RWL                   & \checkmark & 0.1000 & 0.0870 & 0.8300 \\
  Can. + RWL                   & $\times$   & 0.7610 & 0.7540 & 0.5780 \\
  RP                           & \checkmark & 0.1350 & 0.1300 & 0.6910 \\
  RP                           & $\times$   & 0.6450 & 0.6430 & 0.6310 \\
  RP $\rightarrow$ Can.        & \checkmark & 0.0960 & 0.0850 & 0.8430 \\
  RP $\rightarrow$ Can. + RWL  & \checkmark & 0.1010 & 0.0880 & 0.8260 \\
  \bottomrule
\end{tabular}
\vspace{2pt}
\begin{minipage}{\linewidth}
{\footnotesize
  \textbf{Can.}: Canonical ordering.
  \textbf{RWL}: Randomized within levels.\\
  \textbf{RP}: Random permutation per animal.
  \textbf{RN}: Root Normalization.
  \textbf{MPJPE}: Mean Per Joint Position Error.
  \textbf{BE}: Bone Error.
  \textbf{PCP}: Percentage of Correct Parts.
  Lower is better for MPJPE and BE; higher is better for PCP.
}
\end{minipage}
\end{table}

\begin{table}[t]
\centering
\caption{
  Ablation: Unseen (Held-out) Animal Data (Elk)
}\vspace{-1.0em}
\label{tab:ablation_unseen_elk}
\footnotesize
\setlength{\tabcolsep}{4pt}
\begin{tabular}{@{}l c c c c@{}}
  \toprule
  \textbf{Ordering} & \textbf{RN} & \textbf{MPJPE$\downarrow$} & \textbf{BE$\downarrow$} & \textbf{PCP$\uparrow$} \\
  \midrule
  Can.                         & \checkmark & 0.1140 & 0.1020 & 0.8310 \\
  Can.                         & $\times$   & 1.1170 & 1.0990 & 0.6160 \\
  Can. + RWL                   & \checkmark & 0.1190 & 0.1040 & 0.8200 \\
  Can. + RWL                   & $\times$   & 1.0120 & 1.0020 & 0.5410 \\
  RP                           & \checkmark & 0.1746 & 0.1690 & 0.6445 \\
  RP                           & $\times$   & 0.8651 & 0.8658 & 0.5976 \\
  RP $\rightarrow$ Can.        & \checkmark & 0.1166 & 0.1037 & 0.8319 \\
  RP $\rightarrow$ Can. + RWL  & \checkmark & 0.1213 & 0.1057 & 0.8159 \\
  \bottomrule
\end{tabular}
\vspace{2pt}
\vspace{2pt}
\begin{minipage}{\linewidth}
{\footnotesize
  \textbf{Can.}: Canonical ordering.
  \textbf{RWL}: Randomized within levels.\\
  \textbf{RP}: Random permutation per animal.
  \textbf{RN}: Root Normalization.
  \textbf{MPJPE}: Mean Per Joint Position Error.
  \textbf{BE}: Bone Error.
  \textbf{PCP}: Percentage of Correct Parts.
  Lower is better for MPJPE and BE; higher is better for PCP.
}
\end{minipage}\vspace{-1.0em}
\end{table}
The ablation results are reported in Tables~\ref{tab:ablation_study} and~\ref{tab:ablation_unseen_elk}. Root normalization has a large effect across all settings, indicating that cross-skeleton forecasting is sensitive to global translation differences. By expressing poses relative to the root joint, the model can focus on relative body motion rather than absolute scene position. On the full evaluation, random permutation increases MPJPE from 0.094 to 0.135 (43.6\%) and Bone Error from 0.084 to 0.130 (54.8\%), while PCP drops from 0.8413 to 0.691. This shows that matching joint count alone is insufficient; consistent cross-animal joint alignment is also necessary.

\begin{figure}
    \centering
    \includegraphics[width=0.9\linewidth]{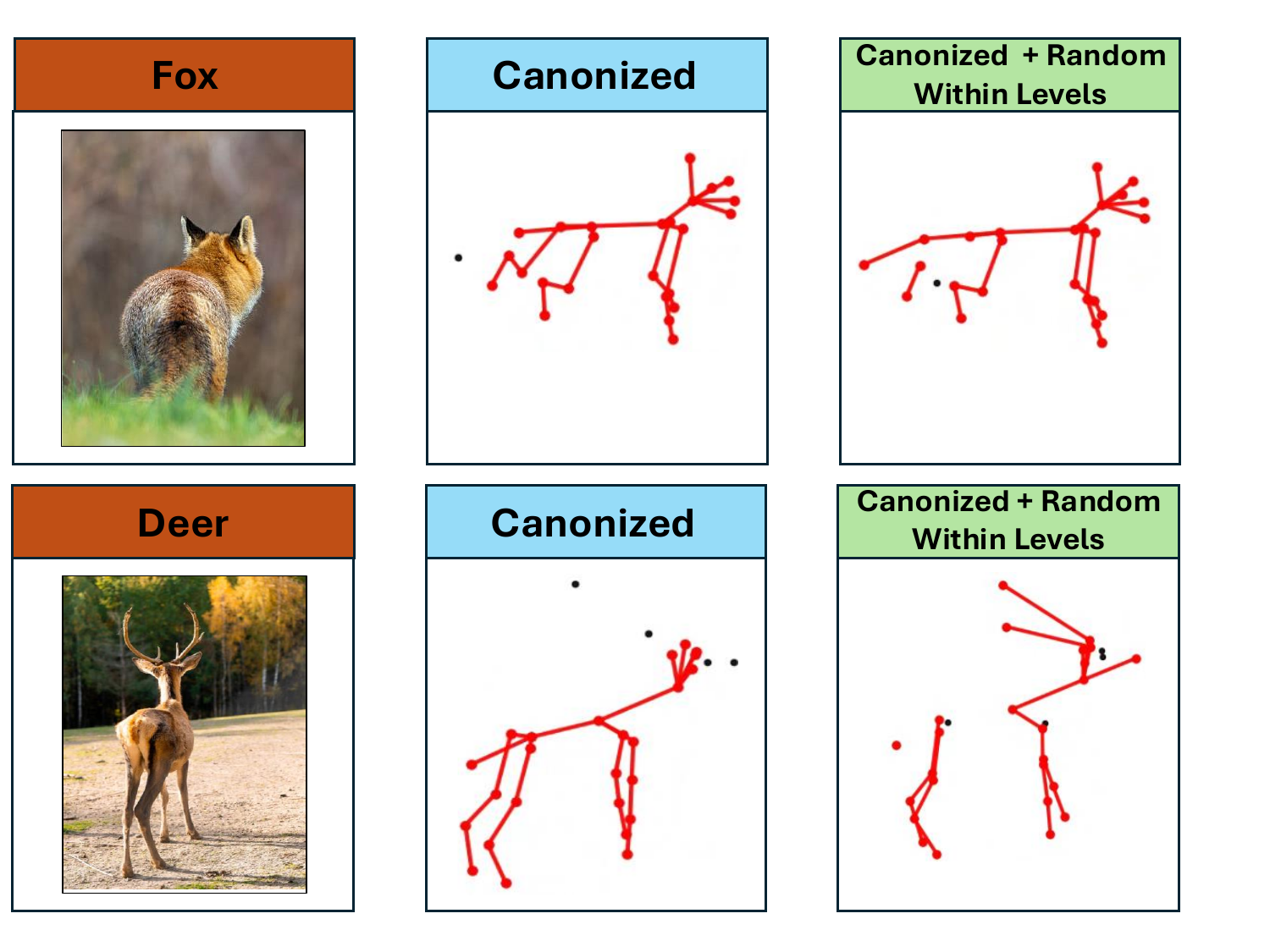}\vspace{-1.0em}
    \caption{Canonical versus random-within-level joint selection for fox and deer skeletons.}\vspace{-2.0em}
    \label{fig:visual_ablation}
\end{figure}
This suggests that canonization is particularly important for unseen animals, where the model cannot rely on animal-specific joint layouts observed during training. Random-within-level variants remain close to the full canonical model but consistently trail it, indicating that coarse BFS-level alignment captures most of the benefit while deterministic within-level selection provides an additional improvement. The following visual ablation further examines how randomized within-level selection affects the semantic and structural consistency of the resulting skeletons.

\subsubsection{Visual Ablation}
We compare fox and deer skeletons under canonical joint selection and random-within-level joint selection. As shown in Figure~\ref{fig:visual_ablation}, canonical selection preserves a shared semantic skeleton across both animals, while random-within-level selection drops structurally important limb joints. Although MPJPE and PCP differ only slightly, the random-within-level skeleton is less semantically consistent and preserves quadruped structure less reliably. Red joints indicate selected joints, while black joints indicate joints dropped to satisfy the shared 25-joint quota.

\section{Discussion and Conclusion}
We presented Semantic-Skeleton, a two-stage framework that maps different quadruped skeletons into a shared joint space before diffusion-based forecasting. By resolving skeletal variation at the representation level, we enable existing forecasting architectures to generalize across species without modification.

Across five forecasting backbones under per-animal skeleton permutations, canonization reduces MPJPE by 49.0\% and improves PCP by 20.8\% on average, with unseen-elk MPJPE reductions reaching 51.0\%. Expanded species-specific baselines further show that Semantic-Skeleton remains competitive with, and in several cases outperforms, models trained directly on fixed native skeletons. On real-world AcinoSet and PFERD data, a single shared model also remains competitive with species-specific models while operating across both skeletons. Ablations identify root normalization, canonical joint ordering, and within-level selection as the key components enabling this generalization. These results show that shared skeletal structure improves cross-species generalization.

\section{Limitations}
Semantic-Skeleton relies on graph topology and is evaluated exclusively on quadrupeds. Despite validation on synthetic and real-world skeletons, generalization to different body plans, such as bipeds, remains untested. Additionally, its heuristic canonization may require manual adaptation for different skeletal structures or datasets. Learned canonization and broader cross-body-plan generalization remain future directions.
{
    \small
    \bibliographystyle{ieeenat_fullname}
    \bibliography{main}
}

\end{document}